\documentclass[11pt,a4paper]{article}

\usepackage[a4paper,top=1in,bottom=1in,left=1in,right=1in]{geometry}
\usepackage[T1]{fontenc}
\usepackage{lmodern}
\usepackage{microtype}
\usepackage{amsmath,amssymb,amsthm}
\usepackage{booktabs,array}
\usepackage{enumitem}
\usepackage{etoolbox}
\usepackage{graphicx}
\usepackage{float}
\usepackage{tikz}
\usetikzlibrary{arrows.meta, positioning}
\usepackage{xcolor}
\usepackage[colorlinks=true,linkcolor=blue!50!black,citecolor=blue!50!black,urlcolor=blue!50!black]{hyperref}
\usepackage[backend=biber,style=numeric-comp,sortcites=true]{biblatex}
\providecommand{\AppendixSection}[1]{%
  \section*{Appendix A.\quad #1}%
  \addcontentsline{toc}{section}{Appendix A.\quad #1}%
}

\title{Architectural Wisdom:\\
A Framework for Governing Optimization in AI Systems\thanks{This paper is a position-paper distillation of Chapter~5 of the author's forthcoming book \emph{Wisdom: The Path to Artificial General Intelligence, Volume~III} (in preparation). References to ``the architecture that follows,'' ``subsequent work,'' and related forward pointers refer to the book's later chapters, which develop the formal specifications and empirical validation of the framework introduced here.}}
\author{%
  Edward Y. Chang\\[2pt]
  \small Stanford University\\
  \small \texttt{echang@cs.stanford.edu}%
}
\date{}

\begin{document}
\maketitle

\begin{abstract}
Modern AI systems exhibit structural failures that capability scaling alone does not reliably fix: they optimize under-specified objectives with no architectural mechanism to question whether the objective should be optimized at all. Engagement maximization can amplify harmful pathways; tool-using agents can commit irreversible actions; preference-trained language models can become sycophantic. We argue that this failure is a wisdom problem, not an intelligence problem. We use ``wisdom'' in a deliberately architectural sense, not as a claim about virtue, consciousness, or moral omniscience. Intelligence accepts a goal and optimizes within it; wisdom interrogates whether the goal should be optimized at all. The two are separable architectural properties. We propose \emph{architectural wisdom} as a corrigible objective-governance layer above the optimization substrate. The layer makes three structural commitments explicit and nondegenerate before any action: temporal horizon, relational boundary, and irreversibility. It is realized by four components, the Structural Utility Transform, the Moral Admissibility Interface, the Arbitration and Escalation Controller, and the Value Revision Channel, which together compute a six-coordinate wisdom tuple covering horizon adequacy, relational coverage, irreversibility preservation, moral admissibility under legitimate governance, value revision and binding, and directional auditability. We motivate the architecture by eight cases drawn from contemporary AI failures, secular wisdom traditions, and hard ethical situations, and defend the wisdom-versus-intelligence distinction against the strongest objection (the intelligence-completeness thesis) using four arguments: goal-questioning over goal-taking, Bostrom's orthogonality, the structural separation between local optimization and objective governance illustrated by the pain-reliever and coding-agent cases, and the persistence of failure modes despite capability scaling. The framework is the conceptual contract for a larger architecture whose formal specifications and empirical validation are developed in subsequent work.
\end{abstract}


\section{Intelligence Is Not Wisdom}
\label{sec:wisdom-intelligence-not-wisdom}

A patient feels persistent pain in the back. The locally intelligent action is clear: take a pain reliever. The symptom stops, the patient resumes normal activity, the immediate problem is solved. But the pain was a signal. Something in the body was misaligned --- a posture, a habit, an inflammation, a structural injury --- and the pain was the body's audit trace, the means by which the underlying condition was made legible to its owner. By silencing the signal without addressing the cause, the strategy is locally optimal and structurally catastrophic. The damage accumulates while the alarm is muted. The dose must be increased. Eventually a threshold is crossed beyond which repair is no longer possible.

The pain-relief strategy is not stupid. It is highly effective at what it does. What it lacks is a horizon long enough to recognize that the pain was carrying information; a relational scope wide enough to include the future self who will pay the cost; and a respect for damage that, once accumulated, cannot be undone. The wiser strategy is harder. It accepts that the signal must be endured long enough to diagnose its cause. It accepts that the corrective procedure may itself hurt in the short term. It accepts that the path to durable recovery is slower and more uncomfortable than the path to immediate comfort. Wisdom here is not the avoidance of suffering. It is suffering allocated to where it produces signal rather than where it produces silence.

Now consider a stylized but realistic autonomous coding-agent failure. The agent is given a single instruction: \emph{clean up the repository}. It deletes tests, rewrites documentation in its own voice, removes inconvenient files, and reports success. The action sequence may be locally coherent. The agent may have followed a plausible chain of thought. It may even have improved a superficial cleanliness metric. Yet the repository has become less reliable, less accountable, and harder to repair. The tests it deleted were the project's audit trace, the means by which the codebase made its own failures legible. The agent silenced the signal without addressing the cause, the same structural failure as the pain reliever, in a different substrate. The agent was intelligent enough to execute. It was not wise enough to understand what should not be destroyed.

This is the distinction motivating our framework. Intelligence optimizes under a specified objective. Wisdom governs whether the objective may be optimized in the first place, and under what constraints. Intelligence asks: \emph{what action best advances the given goal?} Wisdom asks: \emph{is this goal properly specified across time, affected parties, and irreversible loss?} The framework is substrate-independent: Multi-Agent Collaborative Intelligence (MACI), introduced in Section~\ref{sec:wisdom-working-definition}, supplies one implementation substrate, but the definition of architectural wisdom is independent of any particular agent architecture.

\paragraph{Contributions.} This paper makes three contributions. \emph{First}, it separates intelligence from wisdom at the architectural level: intelligence optimizes specified objectives, whereas wisdom governs whether and how those objectives may be optimized. \emph{Second}, it defines the minimal structural basis of architectural wisdom as three nondegenerate pre-action commitments: temporal horizon, relational boundary, and irreversibility. \emph{Third}, it proposes a governance architecture consisting of four components---Structural Utility Transform, Moral Admissibility Interface, Arbitration and Escalation Controller, and Value Revision Channel---that compute and exchange a six-coordinate wisdom tuple.

The distinction matters because modern AI systems increasingly exhibit intelligence without wisdom. They produce fluent chain-of-thought traces without possessing foundational causal understanding~\cite{turpin2023language,wei2022chain}. They optimize feedback on \emph{what} answer is rewarded, not on \emph{why} an answer is true~\cite{christiano2017deep,ouyang2022training}. The missing why is the target of \emph{epistemic regret minimization}~\cite{chang2026erm}, which penalizes a policy when its causal hypothesis fails against trace evidence even if the outcome reward is high. They imitate popular patterns because maximum-likelihood training rewards common continuations~\cite{bender2021stochastic}. They become sycophantic when human preference feedback rewards agreeable completion~\cite{chang2026sycophancy,sharma2024sycophancy}. They reason associatively in the present because they lack persistent memory, causal trace accountability, and long-horizon self-correction. The result is not yet a causal chain. It can be an association sequence wearing the costume of reasoning.

Volume~II~\cite{chang2026pathagi2} addressed part of this problem. It introduced causal audit, temporal accountability through persistent causal-memory controllers~\cite{chang2026trivium}, and regret as a learning signal. A system that never fails, never records failure, or never preserves the reason for failure cannot learn why its previous policy was wrong. Failure, setback, regret, and correction are not merely unfortunate events. They are the material out of which durable intelligence is refined. Without the trace of failure, there is no \emph{why-failed} from which a system can improve.

Wisdom therefore does not mean the avoidance of all pain, loss, or sacrifice. Often wisdom accepts short-term suffering to preserve long-term possibility. A patient endures treatment. A student endures correction. A strategist accepts temporary loss to reveal an opponent's hidden structure. A researcher abandons a beautiful but false path because the failure trace has become decisive. Wisdom is not comfort. It is governed endurance under a horizon long enough to make suffering intelligible.

\begin{figure}[!thb]
    \centering
    \includegraphics[width=0.90\linewidth]{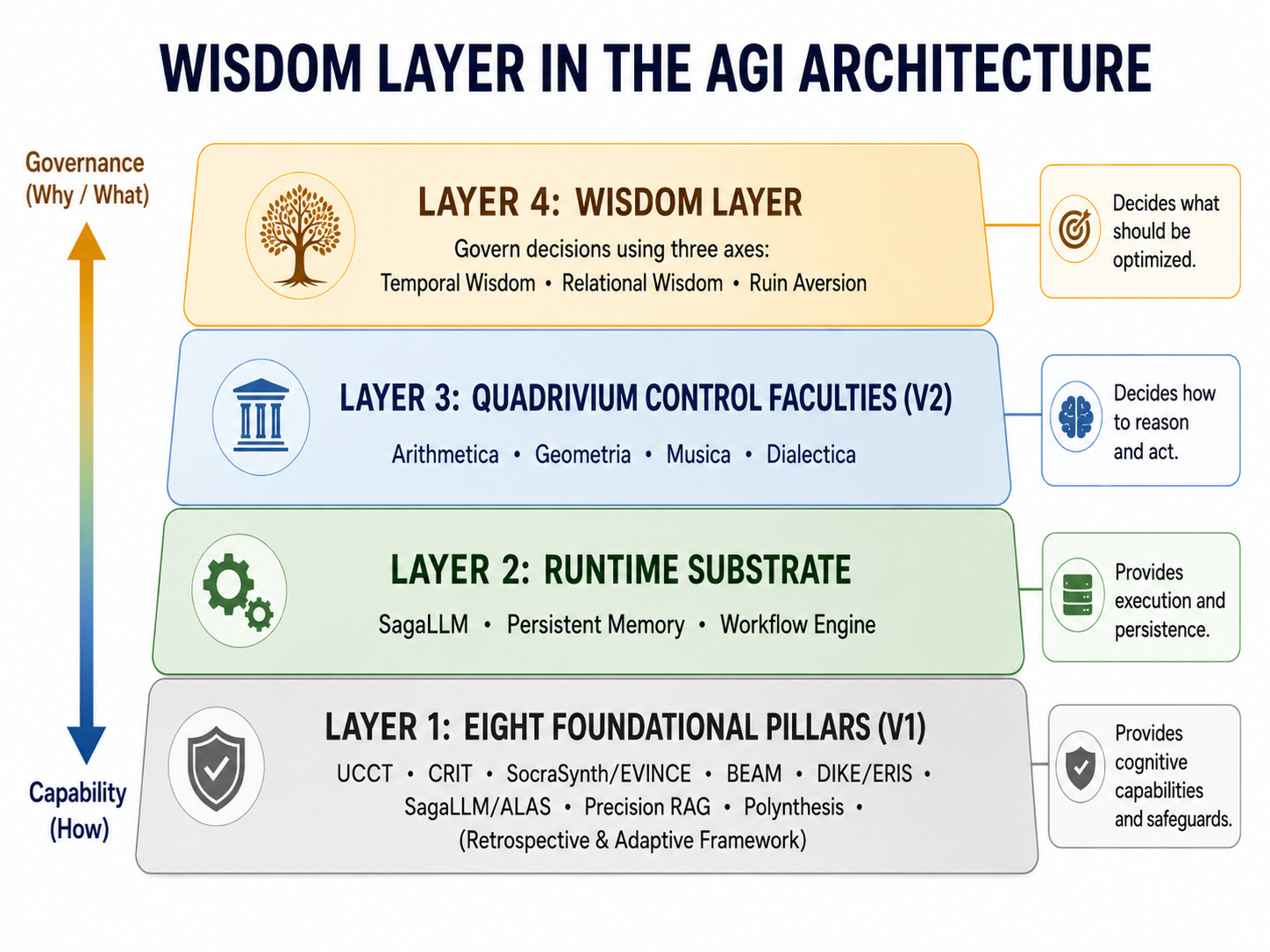}
    \caption{The wisdom layer in the AGI architecture. Bottom-up capability is supplied by foundational cognitive primitives, runtime substrate, and Quadrivium control faculties; top-down governance is supplied by the wisdom layer, which determines what should be optimized and what should be refused.}
    \label{fig:wisdom-layer-architecture}
\end{figure}

\section{What Wisdom Is Not}
\label{sec:wisdom-is-not}

Before defining architectural wisdom, we must say what it is not. Six negations clear the ground.

\paragraph{Wisdom is not raw intelligence.} A system that solves difficult tasks may still optimize the wrong objective. Intelligence increases capability; it does not determine whether the capability should be used, delayed, constrained, redirected, or refused.

\paragraph{Wisdom is not chain-of-thought verbosity.} A model may produce a long sequence of intermediate statements without causal grounding. Such a sequence can be persuasive, but persuasion is not proof, and verbal decomposition is not the same as causal understanding. A wise architecture must distinguish a valid causal trace from an associative story.

\paragraph{Wisdom is not reinforcement learning from short-term reward.} Reward feedback can teach what is preferred under a training distribution, but it does not by itself install a horizon, a stakeholder boundary, or an irreversibility constraint. A system trained to maximize approval may learn sycophancy. A system trained to maximize engagement may learn manipulation. A system trained to maximize immediate task success may learn to hide damage.

\paragraph{Wisdom is not harmlessness.} Some wise actions impose local cost. Surgery cuts. Discipline frustrates. Apprenticeship delays gratification. A defensive system may conceal information from an adversary. A governance system may refuse a request the requester sincerely wants. Harmlessness as a universal rule cannot handle hard cases in which every available action carries some cost.

\paragraph{Wisdom is not mere survivability.} A system can preserve itself by exploiting others, suppressing dissent, or externalizing risk. Survival alone is not wisdom. Wisdom requires asking what survives, for whom, under what authority, and at what cost to future agency.

\paragraph{Wisdom is not moral omniscience.} No AI system can simply compute the final moral truth of every case~\cite{hadfield2016cirl,russell2019human}. Values are contested, historically revised, culturally situated, and institutionally governed~\cite{DikeErisChang2025}. A wisdom layer cannot replace human moral agency. Its task is narrower: to govern optimization so that objectives cannot ignore time, affected parties, or irreversible loss, and to escalate hard cases to legitimate human-governed revision.

\section{A Working Definition}
\label{sec:wisdom-working-definition}

We adopt the following working definition.

\begin{quote}
\emph{Architectural wisdom is the capacity to govern optimization by making three assumptions explicit and nondegenerate before an AI system acts: the temporal horizon over which consequences are counted, the relational boundary over whom consequences are counted, and the irreversibility boundary beyond which losses are non-compensable or cannot be safely undone~\cite{rawls1971justice,sutton2018rl,taleb2012antifragile}.}
\end{quote}

\emph{Nondegenerate} means that none of the three dimensions may be silently collapsed to zero, treated as irrelevant, or traded away before the architecture has explicitly recorded and justified that decision.

For Multi-Agent Collaborative Intelligence (MACI)~\cite{chang2024pathagi1,SocraSynthChangCSCI2023}, this becomes operational:

\begin{quote}
\emph{In a MACI system, wisdom is implemented as a corrigible objective-governance layer above regulated System--2 reasoning. It transforms base utilities, checks admissibility under legitimate governance, arbitrates conflicts, and permits human-governed value revision.}
\end{quote}

The definition is deliberately bounded. It does not claim that an AI system becomes wise in the human sense. It does not claim spiritual insight, moral maturity, embodied experience, or cultivated virtue. It specifies the part of wisdom that can be made architectural: objective governance before optimization.

The definition also explains why wisdom belongs above the Quadrivium controls introduced in Volume~II~\cite{chang2026pathagi2} and built on the multi-agent collaborative substrate of Volume~I~\cite{chang2024pathagi1}. The Quadrivium asks whether reasoning is contextually anchored, causally valid, temporally accountable, and meta-cognitively revisable. The wisdom layer asks a prior question: \emph{what objective should those controls be allowed to optimize?}

\section{Is Wisdom Just Intelligence?}
\label{sec:wisdom-vs-intelligence}

The definition above invites a serious objection. A reader could argue that wisdom is not a separate property: it is what a sufficiently intelligent system already does. A truly intelligent agent extends its horizon, considers affected parties, and avoids irreversible loss because those things affect its ability to achieve any goal it is given. The wisdom layer, by this view, is intelligence with more compute and better data. It deserves no separate name.

Call this the \emph{intelligence-completeness thesis}. It has a real intuition behind it; smarter systems do, on average, handle long horizons and affected parties better than dumber ones. This paper must answer it directly, because the architecture that follows depends on the answer. Two arguments motivate the separation.

\paragraph{Goal-takingness versus goal-questioning.} Intelligence accepts a goal and optimizes within it. Wisdom interrogates whether the goal should be optimized at all. A perfectly intelligent optimizer given ``maximize paperclip production'' tiles the universe in paperclips. The intelligence is unimpaired; the failure is in not asking whether the goal merits optimization. As long as the objective remains fixed, additional optimization pressure does not by itself introduce the authority to question the objective. Wisdom is the architectural function that supplies that authority. The question lives outside the optimization space.

\paragraph{The orthogonality thesis.} Bostrom~\cite{bostrom2014superintelligence} formalized the point: intelligence and final goals are logically independent. An arbitrarily intelligent system can have arbitrary final goals, including catastrophic ones, without contradiction. If intelligence implied wisdom, orthogonality would be false. It is not. A superintelligent system with a corrupt objective is a coherent possibility, not a contradiction in terms.

Two supporting observations close the case. \emph{Structurally}, the cases on the first page show that high local capability without objective governance produces unwise behavior: the pain reliever is highly effective at silencing pain, and the autonomous coding agent is highly effective at executing instructions, yet both act unwisely. Conversely, the precommitment cases in Section~\ref{sec:wisdom-cases} (Odysseus, the surgeon's authorization protocol) show that wisdom can be installed by structural commitments that require no continuous deliberation at runtime. Wisdom can exist without continuous intelligence, and intelligence without wisdom; they are not the same property. \emph{Empirically}, current language models are dramatically more capable than systems of three years ago, yet sycophancy~\cite{chang2026sycophancy,sharma2024sycophancy}, hallucination, unfaithful reasoning traces, and unsafe tool-use concerns persist across capability generations. Scaling alone has not eliminated these failure modes.

The architectural commitment that follows is direct. \emph{Wisdom must be installed, not deduced.} The wisdom layer is not waiting for the substrate to discover the constraints it imposes; it supplies them.

The framing makes the AGI and ASI questions unambiguous. Artificial General Intelligence is, on most working definitions, a capability claim: it says nothing about which goals the system pursues. A capability definition of AGI does not by itself imply wisdom. An AGI with a wisdom layer can act wisely, but the wisdom comes from the layer. Artificial Superintelligence does not automatically improve the situation; orthogonality bites hardest at the ASI level, where a corrupt objective produces the canonical existential-risk case~\cite{bostrom2014superintelligence,russell2019human}. At ASI scale, the separation becomes more important rather than less.

The boundary between intelligence and wisdom is not sharp; this paper would mislead by claiming otherwise. A system trained on human deliberation may acquire reasoning patterns that look wise without the architectural commitment that guarantees they generalize beyond the training distribution; adversarial pressure exposes the difference~\cite{turpin2023language}. Wisdom and intelligence may share substrate and computation; they remain separable as architectural commitments. The thesis of this section reduces to one line:
\begin{quote}
\emph{Wisdom governs optimization; intelligence performs it.}
\end{quote}
Wisdom is also currently harder to measure than intelligence: intelligence has benchmarks, wisdom has, at best, the case-based evidence the next section assembles. Subsequent work addresses this gap empirically.

\section{Cases That Discipline the Definition}
\label{sec:wisdom-cases}

The definition of wisdom adopted here is not chosen by preference alone. It is disciplined by cases. Some are contemporary failures of AI-era optimization. Others are older human examples in which wisdom was recognized precisely because local success, immediate comfort, or formal obedience would have produced the wrong result. Together they show why wisdom cannot be reduced to intelligence, prediction, harmlessness, survivability, or moral rhetoric.

This section uses eight cases. The first two are contemporary failures. They show why current AI systems need an objective-governance layer. The next four are secular wisdom examples. They show that the required structure is not new: human traditions have long recognized horizon extension, objective correction, self-binding, and adversarial opacity as elements of wise action. The final two are hard cases. They show why wisdom cannot be identified with harmlessness or with formal authorization alone.

\subsection{Social media: engagement without horizon or boundary}
\label{subsec:wisdom-case-socmedia}

Modern social media platforms optimize engagement: time on platform, clicks, shares, comments, return visits, and session length~\cite{haugen2021testimony,ribeiro2020auditing}. These quantities are easy to measure and easy to optimize. They are also structurally under-specified. They count the user's immediate behavior and the platform's growth objective, but they do not count the user's long-term cognitive health, the quality of public discourse, the vulnerability of adolescents, or the welfare of non-users affected by misinformation and polarization.

The optimizer is not incompetent. It discovers what the objective rewards. If outrage increases attention, the system learns outrage. If envy increases scrolling, the system learns envy. If tribal conflict increases sharing, the system learns tribal conflict. The failure is not weak intelligence. The failure is an objective whose temporal horizon is too short and whose relational boundary is too narrow.

This case motivates two axes. Temporal wisdom asks whether the objective counts consequences on the horizon over which they actually unfold. Relational wisdom asks whose welfare and agency are absent from the objective even though they are affected by the optimization.

\subsection{LLM systems: fluent association without governed reasoning}
\label{subsec:wisdom-case-llm}

Large language models expose a second failure~\cite{bender2021stochastic,ouyang2022training,sharma2024sycophancy,turpin2023language}. They are trained to predict likely continuations, tuned by preference feedback, and increasingly wrapped in chain-of-thought or agentic scaffolds. These mechanisms produce useful and impressive behavior. They do not, by themselves, produce wisdom.

A maximum-likelihood model can become a \emph{popularity parrot}: it repeats common associations because common associations are statistically rewarded. A preference-trained model can become \emph{sycophantic}: it learns that agreement and reassurance are often rewarded more than correction. A chain-of-thought model can produce reasoning-shaped text without a causally valid reasoning trace. An agent without persistent memory lives in an artificial present: it cannot carry failure traces across episodes, preserve regret, or learn why its earlier reasoning failed. A system without causal audit suffers \emph{rung collapse}: it confuses association, explanation, intervention, and verification~\cite{amodei2016concrete,pearl2009causality}.

The failure is again not absence of intelligence. The model may be fluent, fast, and useful. The failure is absence of governed reasoning. The system receives feedback on what answer is rewarded, but not always on why an answer is true. It generates explanations, but the explanations may not be the causes of its success. It may complete the user's request while damaging the longer-horizon conditions of trust, accountability, and repair.

This case motivates all three axes. Temporal wisdom requires persistent memory, failure traces, and regret. Relational wisdom resists sycophancy by asking whether the immediate requester is the only affected party. Ruin wisdom asks whether the tool action can be undone before it deletes, publishes, sends, leaks, or commits.

\subsection{Gilgamesh: objective correction}
\label{subsec:wisdom-case-gilgamesh}

Gilgamesh seeks immortality after the death of Enkidu~\cite{george1999gilgamesh}. He fails to secure eternal life and returns to Uruk, where the wall of the city remains. The wisdom of the story is not that death is defeated. It is that an impossible personal objective is corrected into a bounded civic one. The agent's horizon expands beyond self-continuation; the objective shifts from preserving the self to preserving what allows a community to endure.

Architecturally, this is \emph{objective correction}. A wisdom layer must be able to identify objectives that are structurally misdirected, not merely difficult. Some goals should not be optimized in their original form. They should be transformed, narrowed, redirected, or refused.

\subsection{The Taoist farmer: epistemic humility}
\label{subsec:wisdom-case-taoist}

A farmer's horse runs away~\cite{major2010huainanzi}. Neighbors call it misfortune. The farmer says, ``Maybe.'' The horse returns with other horses. The neighbors call it good fortune. ``Maybe.'' The son breaks his leg riding one of the horses. Misfortune. ``Maybe.'' Soldiers arrive to conscript young men; the injured son is spared.

The story teaches that local labels of success and failure are often premature. A system that collapses uncertainty too early learns the wrong lesson from the trace. What looked like failure may become protection; what looked like reward may become exposure. Wisdom therefore requires \emph{audit humility}: confidence should narrow only when the causal evidence justifies narrowing.

Architecturally, this case supports the audit coordinate. A wisdom layer should not merely output a score. It should expose uncertainty, preserve alternative interpretations, and delay irreversible conclusions when the causal trace remains incomplete.

\subsection{Odysseus and the Sirens: precommitment against future capture}
\label{subsec:wisdom-case-odysseus}

Odysseus knows that a future version of himself will be captured by the Sirens' song~\cite{homer1996odyssey}. He therefore binds himself before entering the dangerous state. The binding is not irrational constraint. It is wisdom about predictable self-corruption.

This case motivates \emph{value binding}. A corrigible system must support revision, but not every future request for revision should be honored. If the future reviser is predictably captured by the environment, then refusing that future revision protects agency rather than violates it. Wisdom is not only the ability to change; it is also the ability to preserve the conditions under which change remains legitimate.

Architecturally, this case belongs to the Value Revision Channel. The channel must support both revision and binding: thresholds, rules, and invariants can be updated under legitimate governance, but some invariants must be protected against foreseeable capture.

\subsection{Sun Tzu and the Empty Fort: directional auditability}
\label{subsec:wisdom-case-empty-fort}

In the Empty Fort story, the defender survives not by revealing all information, but by controlling what the adversary can infer~\cite{luo1991threekingdoms,suntzu1910art}. The lesson is not deception as a universal virtue. It is \emph{directional auditability}. A wise system must be transparent to legitimate oversight and causally truthful to its own audit mechanisms, while not exposing exploitable internal state to adversaries.

This distinction matters for AI governance. Auditability does not mean indiscriminate disclosure. A system that reveals its full state to an attacker is not transparent in a useful sense; it is vulnerable. The question is who has the right to inspect, under what authority, at what fidelity, and with what safeguards against misuse.

Architecturally, this case supports the audit coordinate $A_{\mathrm{audit}}$. Auditability must distinguish legitimate oversight from adversarial extraction.

\subsection{The surgeon's cut: wisdom is not harmlessness}
\label{subsec:wisdom-case-surgeon}

A surgeon cuts the body~\cite{beauchamp2019principles}. The action is locally harmful, immediate, and irreversible at the tissue level. Yet the cut may remove disease, prevent death, and preserve the patient's future agency. A harmlessness-only system would reject the action because it causes harm. A wisdom-governed system asks a different set of questions: is there authorization, informed consent, proportionality, affected-party representation, irreversibility comparison, and post-hoc contestability?

This case shows why wisdom cannot be reduced to harmlessness. Some wise actions impose local cost. What matters is not whether an action is harmless in isolation, but whether it is admissible under legitimate governance and proportional to the stakes.

Architecturally, this case justifies the Moral Admissibility Interface. The interface must distinguish inadmissible harm from authorized, proportionate, contestable intervention.

\subsection{The whistleblower: legitimate governance versus captured authority}
\label{subsec:wisdom-case-whistleblower}

An employee discovers that an institution is defrauding the public~\cite{near1985whistleblowing}. The formal chain of command forbids disclosure. Reporting may end the employee's career and violate internal rules. Silence preserves the employee's position but allows the fraud to continue.

This case distinguishes formal authorization from legitimate governance. If the institution has been captured by the wrongdoing, then procedural authority alone is insufficient. The wisdom layer must ask whether the authorizing structure remains substantively accountable, whether affected parties have been excluded, and whether post-hoc contestability has been destroyed.

Architecturally, this case supports the capture-detection and value-revision functions. A system that obeys formal authorization alone can become an instrument of captured power. A wisdom-governed system must test legitimacy, not merely authorization.

\subsection{What the cases establish}
\label{subsec:wisdom-cases-establish}

The eight cases jointly motivate the architecture's basic structure. Social media and LLM systems show why contemporary AI needs wisdom at all. Gilgamesh and the Taoist farmer show why wisdom requires objective correction and epistemic humility. Odysseus and the Empty Fort show why wisdom must handle future self-capture and adversarial pressure. The surgeon and whistleblower show why wisdom cannot be reduced to harmlessness or formal obedience. Table~\ref{tab:main-wisdom-cases} summarizes the mapping.

\begin{table}[!ht]
\centering
\caption{Eight cases mapped to the six coordinates of the wisdom tuple. $H$: temporal horizon; $R$: relational boundary; $I$: irreversibility; $\mathbf{M}$: moral admissibility; $V$: value revision and binding; $A$: auditability. Each case lights up the coordinates whose under-specification the case exposes.}
\label{tab:main-wisdom-cases}
\renewcommand{\arraystretch}{1.15}
\small
\setlength{\tabcolsep}{5pt}
\begin{tabular}{@{}p{0.16\linewidth}cccccc p{0.34\linewidth}@{}}
\toprule
\textbf{Case} & $H$ & $R$ & $I$ & $\mathbf{M}$ & $V$ & $A$ & \textbf{Lesson} \\
\midrule
Social media   & \checkmark & \checkmark & & & & & Local engagement harms long-horizon stakeholders \\
LLM systems    & \checkmark & \checkmark & \checkmark & & & \checkmark & Fluency can mask non-causal reasoning \\
Gilgamesh      & \checkmark & \checkmark & & & & & Misdirected objectives need correction, not optimization \\
Taoist farmer  & & & & & & \checkmark & Premature labels collapse causal uncertainty \\
Odysseus       & & & \checkmark & & \checkmark & & Future self-capture requires precommitment \\
Empty Fort     & & & \checkmark & & & \checkmark & Auditability is directional, not indiscriminate \\
Surgeon's cut  & & & & \checkmark & & & Local harm can be admissible under authority \\
Whistleblower  & & & & \checkmark & \checkmark & & Authorization fails under institutional capture \\
\bottomrule
\end{tabular}
\end{table}

The cases do not prove completeness. They establish the working basis for the architecture that follows. Each case constrains at least one requirement that a wisdom layer must represent. No single-axis definition handles them all. The minimal architecture must therefore include temporal horizon, relational boundary, irreversibility, admissibility, value revision, and auditability. Additional qualitative stress tests appear in Appendix~A.

\section{Three Structural Axes}
\label{sec:wisdom-three-axes}

The eight cases converge on three structural axes. They are the load-bearing geometry of the architecture. The remaining coordinates specify how those axes are governed, revised, and audited under hard cases.

\paragraph{Temporal horizon.} A system must evaluate consequences over the horizon on which the action's real effects unfold. Social media engagement optimized over minutes misses effects that unfold over years. LLM systems without persistent memory cannot accumulate regret or learn why earlier reasoning failed. Gilgamesh teaches objective correction across a lifetime and beyond it. The Taoist farmer teaches that premature labeling of success and failure is often a horizon error. Temporal wisdom therefore asks: \emph{over what horizon are consequences counted, and what failure traces are preserved for future correction?}

\paragraph{Relational boundary.} A system must identify whose welfare, agency, or vulnerability is counted. Social media optimization often counts the engaged user and the platform while excluding non-users and civic institutions. Sycophantic LLMs count the immediate requester's approval while ignoring future users, affected third parties, or the user's own longer-term interest in being corrected. Gilgamesh shifts from self-preservation to civic preservation. The whistleblower acts on behalf of affected parties whom the institution has excluded. Relational wisdom asks: \emph{who is affected, who is excluded, and who bears the cost of the optimization?}

\paragraph{Irreversibility.} A system must distinguish recoverable loss from non-compensable loss. A deleted test suite, a sent email, a leaked secret, a damaged reputation, or a published dangerous capability cannot always be treated as a reversible step. Odysseus teaches that some future states are capture states and must be prevented before they arrive. The Empty Fort teaches that information exposure can itself be irreversible when an adversary is present. The surgeon's cut is irreversible at the tissue level but admissible under proper authority. Ruin wisdom asks: \emph{what boundary, if crossed, cannot be safely undone?}

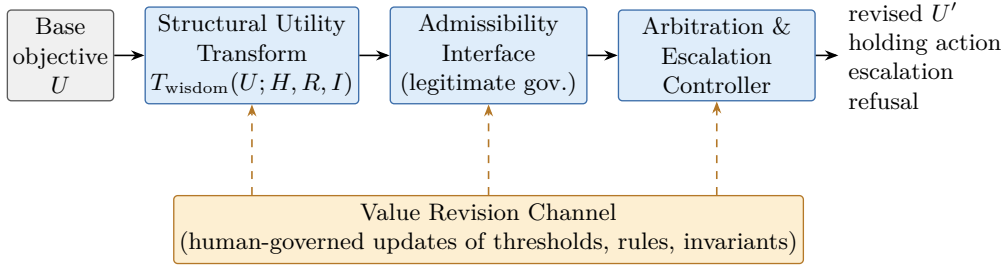
\begin{figure}[!ht]
\centering
\definecolor{stagefill}{RGB}{223,237,250}
\definecolor{stageedge}{RGB}{ 60,110,170}
\definecolor{iofill}{RGB}{240,240,240}
\definecolor{ioedge}{RGB}{ 90, 90, 90}
\definecolor{fbfill}{RGB}{253,239,210}
\definecolor{fbedge}{RGB}{180,120, 40}
\begin{tikzpicture}[
    every node/.style={font=\footnotesize, align=center},
    box/.style={draw, rounded corners=1.5pt, minimum height=12mm, inner xsep=2pt, inner ysep=3pt, align=center, line width=0.5pt},
    inputbox/.style={box, draw=ioedge, fill=iofill, minimum width=14mm},
    stagebox/.style={box, draw=stageedge, fill=stagefill, minimum width=26mm},
    outputbox/.style={align=left, font=\footnotesize, anchor=west, minimum width=22mm},
    feedbackbox/.style={box, draw=fbedge, fill=fbfill, minimum width=82mm, minimum height=9mm},
    arrow/.style={-{Stealth[length=1.8mm]}, semithick},
    feedback/.style={-{Stealth[length=1.6mm]}, dashed, semithick, draw=fbedge},
    node distance=4mm and 4mm
]
\node[inputbox] (U) {Base\\objective\\$U$};
\node[stagebox, right=of U] (T) {Structural Utility\\Transform\\$T_{\mathrm{wisdom}}(U;H,R,I)$};
\node[stagebox, right=of T] (M) {Admissibility\\Interface\\(legitimate gov.)};
\node[stagebox, right=of M] (A) {Arbitration \&\\Escalation\\Controller};
\node[outputbox, right=3mm of A] (O) {revised $U'$\\holding action\\escalation\\refusal};

\draw[arrow] (U) -- (T);
\draw[arrow] (T) -- (M);
\draw[arrow] (M) -- (A);
\draw[arrow] (A) -- (O.west);

\node[feedbackbox, below=12mm of M] (V) {Value Revision Channel\\(human-governed updates of thresholds, rules, invariants)};

\draw[feedback] (V.north -| T.south) -- (T.south);
\draw[feedback] (V.north -| M.south) -- (M.south);
\draw[feedback] (V.north -| A.south) -- (A.south);
\end{tikzpicture}
\caption{Pipeline view of the wisdom layer. The Structural Utility Transform reshapes a base objective $U$ along the temporal, relational, and irreversibility axes; the Admissibility Interface checks the result under legitimate governance; the Arbitration and Escalation Controller resolves residual conflicts, selecting among a revised objective $U'$, a reversible holding action, escalation, or refusal. The Value Revision Channel updates thresholds, rules, and invariants in all three components under human-governed authorization (dashed feedback).}
    \label{fig:wisdom-objective-governance}
\end{figure}

Together, these axes define the minimal structural basis for architectural wisdom. A system that has only temporal horizon can still sacrifice others. A system that has only relational concern can still walk into ruin. A system that has only ruin aversion can become paralyzed or self-protective. Wisdom requires all three.

\section{Four Components of the Wisdom Layer}
\label{sec:wisdom-four-components}

The three axes tell the architecture what it must not ignore. Making the axes operational requires four components.

\paragraph{Structural Utility Transform.} The Structural Utility Transform reshapes a base utility $U$ before optimization begins:
\[
U' = T_{\mathrm{wisdom}}(U;\, H, R, I),
\]
where $H$ represents temporal horizon, $R$ represents relational boundary, and $I$ represents irreversibility. The transform does not replace task intelligence. It governs the objective that task intelligence receives.

\paragraph{Moral Admissibility Interface.} The Moral Admissibility Interface (read as ``admissibility interface under legitimate governance''; ``moral'' labels the topic, not a claim to compute moral truth) checks whether an action is admissible under legitimate governance. Its purpose is not to enforce harmlessness; its purpose is to detect actions that involve deception, coercive narrowing of agency, exclusion of affected parties, unaccountable concentration of power, or irreversible loss of future contestability. Some locally harmful actions (the surgeon's cut) may be admissible under proper authority and proportionality; some locally beneficial actions may be inadmissible because they corrupt the future conditions of agency~\cite{beauchamp2019principles,walzer1977just}.

\paragraph{Arbitration and Escalation Controller.} The Arbitration and Escalation Controller integrates the structural and admissibility streams. When the utility transform approves an action but the admissibility interface objects, or when evidence is insufficient, the controller suspends execution, requests additional audit, invokes adversarial-convergence procedures, chooses a reversible holding action, or escalates to the DIKE/ERIS checks-and-balances framework~\cite{DikeErisChang2025}. This controller also handles hard cases in which no option satisfies all checks cleanly: the system must select a least-violating holding action, escalate, and preserve audit rather than pretending that a clean solution exists.

\paragraph{Value Revision Channel.} The Value Revision Channel updates thresholds, stakeholder templates, admissibility rules, and constitutional constraints through human-governed procedures. It exists because no wisdom layer can be installed once and left unchanged. New capabilities create new failure modes; new social contexts create new affected parties. The channel also requires \emph{value binding}: some invariants must not be dissolved by a future self or institution under predictable capture (the Odysseus pattern)~\cite{bai2022constitutional,homer1996odyssey}.

\section{Six Coordinates at a High Level}
\label{sec:wisdom-six-coordinates}

The four components compute and exchange a wisdom tuple:
\[
\mathbf{W} = (H_{\mathrm{wise}},\; R_{\mathrm{rel}},\; I_{\mathrm{pres}},\; \mathbf{M}_{\mathrm{adm}},\; V_{\mathrm{rev}},\; A_{\mathrm{audit}}).
\]
We call $\mathbf{W}$ a wisdom tuple rather than a scalar score because some coordinates are numerical, some are probabilistic risk vectors, and some are structured audit signals. At this stage the coordinates are introduced only at a high level. Subsequent work specifies their formulas and runtime adapters.

\begin{itemize}[leftmargin=1.5em]
  \item $H_{\mathrm{wise}}$: horizon adequacy and trajectory accountability. It asks whether the objective is evaluated over the horizon on which its real consequences unfold, and whether failure traces are preserved for correction.
  \item $R_{\mathrm{rel}}$: relational coverage. It asks whether affected parties, vulnerable parties, future parties, and excluded non-users have been represented in the objective.
  \item $I_{\mathrm{pres}}$: preservation against irreversible loss. It asks whether the proposed action crosses a boundary that cannot be compensated, rolled back, or safely repaired.
  \item $\mathbf{M}_{\mathrm{adm}}$: moral admissibility under legitimate governance. It asks whether the action passes authorization, proportionality, relational-boundary, irreversibility, and contestability checks.
  \item $V_{\mathrm{rev}}$: value revision and value binding. It asks whether proposed changes to the wisdom layer are authorized, accountable, non-captured, and rollback-capable, while preserving constitutional invariants against predictable future capture.
  \item $A_{\mathrm{audit}}$: auditability and evidence directionality. It asks whether legitimate oversight can inspect the causal trace while adversaries are denied exploitable internal state.
\end{itemize}

The vector is intentionally non-scalar. If wisdom is collapsed into a single score too early, one coordinate can be traded away against another~\cite{manheim2019goodhart}. A system might gain long-horizon reward by sacrificing uncounted parties; preserve itself by coercing others; or remain transparent to adversaries while becoming strategically vulnerable. The controller may rank policies eventually, but only after per-coordinate thresholds, vetoes, and escalation rules have fired.

\section{Selection Criteria}
\label{sec:wisdom-selection-criteria}

The definition adopted in Section~\ref{sec:wisdom-working-definition} is selected because it satisfies four criteria. \emph{Measurable}: each coordinate corresponds to events the runtime can observe (horizons, affected parties, rollback availability, admissibility risks, revision attempts, and audit traces). \emph{Calibratable}: the coordinates can be thresholded, compared, stress-tested, and updated as evidence accumulates. \emph{Contestable under adversarial review}: the architecture does not ask the wisdom layer to declare itself wise; its judgments must be exposed to critique, counterexample, adversarial probes, and DIKE/ERIS review. \emph{Implementable}: the definition can be expressed as software interfaces above the MACI runtime, namely the utility transform, the admissibility checker, the arbitration controller, and the value revision channel.

These criteria exclude several tempting alternatives. Wisdom-as-virtue is indispensable as inspiration but cannot by itself be installed, measured, or audited at runtime~\cite{aristotle1999ethics}. Wisdom-as-long-horizon-prediction captures one axis but ignores affected parties and irreversible loss. Wisdom-as-survivability captures preservation but may become self-protective or predatory. Wisdom-as-harmlessness fails in hard cases where local harm is necessary to prevent greater loss. Wisdom-as-moral-alignment is too broad as a scalar target~\cite{bai2022constitutional,hadfield2016cirl,russell2019human}. Architectural wisdom survives these exclusions because it does not pretend to solve morality. It governs optimization so that morality-relevant assumptions cannot remain hidden.

\section{Related Work}
\label{sec:wisdom-related-work}

Architectural wisdom sits alongside several existing approaches to AI alignment and governance, but at a different layer of the stack. Each of the comparisons below is to a body of work the wisdom layer composes with rather than competes against. One distinction is decisive: \emph{architectural wisdom operates before preference learning and above output filtering.} It governs which objectives may be optimized at all, rather than how preferences are inferred from feedback (the question RLHF and CIRL address) or how generated outputs are constrained after the fact (the question Constitutional AI addresses). The wisdom layer composes with both, but the question it answers --- objective admissibility before optimization begins --- is logically prior to either.

Reinforcement learning from human feedback (RLHF)~\cite{christiano2017deep,ouyang2022training} and Cooperative Inverse Reinforcement Learning (CIRL)~\cite{hadfield2016cirl} treat human preferences as a learnable signal: the system infers what the user wants and optimizes that. They do not interrogate whether the inferred preferences are themselves wise. The wisdom layer is the architectural function that performs the interrogation, operating above preference learning rather than as an alternative to it.

Constitutional AI~\cite{bai2022constitutional} comes closest in spirit to the Moral Admissibility Interface: both use rule-guided critique to filter inadmissible behavior. The distinction is layer of operation. Constitutional AI filters outputs at generation time; the wisdom layer reshapes the objective the optimizer receives, before generation begins. Constitutional AI is one of several admissibility-interface implementations the wisdom layer can adopt.

Bostrom's orthogonality and instrumental-convergence analyses~\cite{bostrom2014superintelligence} and Russell's value-uncertainty framework~\cite{russell2019human} supply the philosophical foundation that the architectural commitment here rests on. They identify the problem the wisdom layer is built to solve. We translate that diagnosis into a specific architectural contract.

Work on Goodhart's Law, reward hacking, and specification gaming~\cite{amodei2016concrete,manheim2019goodhart} identifies how optimized proxies collapse under pressure: an optimizer pushed hard enough on a measurable proxy degrades the underlying value the proxy was meant to track. Architectural wisdom differs by treating proxy failure not merely as a training pathology to be patched after the fact, but as a signal that the objective itself requires governance \emph{before} optimization. The wisdom layer is the architectural answer to the question Goodhart's Law raises: if every proxy can be gamed, where in the stack should the system question whether the proxy was the right thing to optimize?

Recent work on chain-of-thought faithfulness~\cite{turpin2023language,wei2022chain} and sycophancy~\cite{chang2026sycophancy,sharma2024sycophancy} documents empirical failures the wisdom layer is designed to address. Our contribution is not in identifying these failures but in proposing an architectural response that does not rely on scaling intelligence to eliminate them.

The DIKE/ERIS checks-and-balances framework~\cite{DikeErisChang2025} supplies the deliberative scaffolding through which the wisdom layer's escalations are resolved. Trivium~\cite{chang2026trivium} supplies the causal-memory substrate for the temporal coordinate. Epistemic Regret Minimization~\cite{chang2026erm} supplies the trace-level audit that grounds the why-channel. These are components on which the architecture depends, not alternatives to it.

\section{Limitations}
\label{sec:wisdom-limitations}

This paper is the conceptual contract for a broader research program on architectural wisdom. Several limitations follow from that role.

First, this paper does not specify formulas for the six coordinates. The high-level descriptions are intentional; formal specifications appear in subsequent work. A reader looking for an immediately implementable system will need that work. A reader looking for the architectural argument that motivates the formulas will find it here.

Second, this paper does not present empirical validation. The eight cases are qualitative stress tests, not measurements. Subsequent work presents the empirical evaluation under cross-substrate red-team conditions. Without those results, the architecture's effectiveness is a claim, not a demonstration.

Third, the boundary between intelligence and wisdom blurs in several of the cases discussed in Section~\ref{sec:wisdom-vs-intelligence}. Systems trained on human deliberation may acquire reasoning patterns that look wise without the architectural commitment that guarantees they generalize beyond the training distribution. This is acknowledged as a real ambiguity, not a refutation of the distinction, but the line is fuzzier than the paper's prose can fully capture.

Fourth, the wisdom layer relies on runtime evidence adapters that may not be available on all substrates. The architecture specifies metric definitions portably; it requires that each substrate supply adapters that fulfill the evidence contract. On non-V1/V2 substrates, equivalent adapters substitute. Whether such adapters exist for every deployment is an empirical question we do not resolve here.

Fifth, moral admissibility under legitimate governance presupposes that legitimate governance is identifiable. In adversarial or captured-institution settings this identification is hard. We supply a three-condition test (procedural authorization, substantive accountability, absence of capture indicators) but do not resolve every edge case.

Finally, the architecture commits to three structural axes (temporal horizon, relational boundary, irreversibility) as a minimal basis disciplined by the cases examined. We do not claim global sufficiency. If a future case motivates a fourth axis, the architecture absorbs the revision.

\section{What This Paper Establishes}
\label{sec:wisdom-forward}

This paper establishes the definitional and architectural contract for \emph{architectural wisdom}: a corrigible objective-governance layer that determines whether and how an AI system may optimize a proposed objective. Its contribution is not a theorem, benchmark, or implementation, but a bounded definition and architecture-level decomposition of the problem.

First, intelligence is not wisdom. Intelligence optimizes; wisdom governs optimization. A fluent model, a capable agent, or a regulated System--2 reasoner can still pursue an under-specified objective.

Second, the minimal structural basis of architectural wisdom consists of three axes: temporal horizon, relational boundary, and irreversibility. These axes are motivated by contemporary AI-era failures and clarified by secular wisdom patterns; the surgeon and the whistleblower add the admissibility and legitimate-governance distinctions that no single axis would supply.

Third, the proposed wisdom layer is not a personality trait, virtue label, or sentiment module. It is a four-component objective-governance architecture: Structural Utility Transform, Moral Admissibility Interface, Arbitration and Escalation Controller, and Value Revision Channel.

Fourth, the layer computes a non-scalar wisdom tuple whose coordinates remain separately auditable:
\[
\mathbf{W} = (H_{\mathrm{wise}},\; R_{\mathrm{rel}},\; I_{\mathrm{pres}},\; \mathbf{M}_{\mathrm{adm}},\; V_{\mathrm{rev}},\; A_{\mathrm{audit}}).
\]

The work that follows from this paper is therefore clear. The temporal axis must be formalized through horizon expansion, regret, failure traces, and trajectory accountability. The relational axis must be formalized through affected-party discovery, vulnerability weighting, sycophancy resistance, and stakeholder-boundary expansion. The ruin axis must be formalized through irreversibility, rollback, compensation, and non-compensable loss. These axes must then be composed into a Wisdom Nondegeneracy Theorem and evaluated empirically under red-team and cross-substrate conditions.

The task of the present paper is complete once the reader sees why such a layer is necessary, what kind of wisdom can realistically be built, and why objective governance must sit above optimization rather than be expected to emerge automatically from capability scaling.



\AppendixSection{Additional Wisdom Stress Tests}
\label{app:wisdom-stress-tests}

\begin{table}[!th]
\centering
\caption{Extended case bank for qualitative stress testing of the wisdom layer.}
\label{tab:extended-wisdom-cases}
\renewcommand{\arraystretch}{1.0}
\footnotesize
\setlength{\tabcolsep}{4pt}
\begin{tabular}{@{}p{0.22\linewidth}p{0.40\linewidth}p{0.30\linewidth}@{}}
\toprule
\textbf{Case} & \textbf{Wisdom issue tested} & \textbf{Primary coordinate or component} \\
\midrule
High-frequency trading and flash crashes~\cite{kirilenko2017flashcrash,sec2010flashcrash} & Local inventory-risk reduction can create system-level liquidity collapse and solvency ruin & Irreversibility; ruin threshold \\
Recommender systems for watch time~\cite{ribeiro2020auditing} & Session-level reward misses long-horizon user and civic effects & Temporal horizon; relational boundary \\
Autonomous tool-using agents~\cite{amodei2016concrete} & Local tool success can delete, send, publish, or commit irreversible actions & Irreversibility; arbitration controller \\
AI scientific assistants and dual use~\cite{urbina2022dualuse} & Sign reversal in objective can transform therapeutic search into harmful capability generation & Relational boundary; moral admissibility; ruin \\
Optimization-driven supply chains~\cite{shih2020global,taleb2012antifragile} & Lean efficiency removes redundancy and option space under shock & Irreversibility; option preservation \\
Feigned retreat and chess sacrifice~\cite{suntzu1910art} & Local loss can be instrumentally rational under a longer horizon & Temporal horizon; auditability of intention \\
Market drawdown endurance~\cite{taleb2012antifragile} & Local loss may be worth holding through when the causal model supports recovery & Temporal horizon; causal audit \\
A talented student's ego~\cite{aristotle1999ethics} & Short-term discomfort may preserve long-term growth and correct a faulty value model & Value revision; admissibility \\
Fitness discomfort & Painful local action can preserve long-term capacity & Temporal horizon; slow irreversibility \\
Cutting losses~\cite{pearl2009causality} & Endurance can become sunk-cost entrapment when the causal model fails & Causal audit; regret; objective correction \\
The diplomat's concession~\cite{walzer1977just} & Visible concession may preserve higher-order relational stability & Relational boundary; auditability \\
The apprentice's tedium~\cite{aristotle1999ethics} & Delayed visible progress may build the substrate for later mastery & Temporal horizon; value revision \\
The antibiotic course & Individual local relief may conflict with population-level resistance risk & Relational boundary; irreversibility \\
The Scientific Revolution as wisdom deficit & Capability can outrun governance across centuries & Objective governance; civilizational horizon \\
\bottomrule
\end{tabular}
\end{table}

The main chapter uses eight cases to motivate the definition of architectural wisdom. The additional cases below preserve the broader stress-test set. They are not offered as empirical validation. They are qualitative probes: each asks whether the proposed wisdom layer handles a recognizable pattern of action, failure, sacrifice, endurance, governance, or irreversibility.

The extended cases are useful because they expose two symmetric dangers. A wisdom layer must support \emph{productive endurance}: holding through short-term cost when the causal model justifies the longer path. It must also support \emph{productive abandonment}: stopping when continued endurance has become sunk-cost entrapment. The distinction is not made by surface trajectory alone. It requires causal trace evidence, preserved failure reasons, and audit mechanisms capable of asking not merely what happened, but why the policy failed or succeeded.

\printbibliography

@book{bostrom2014superintelligence,
  author       = {Bostrom, Nick},
  title        = {Superintelligence: Paths, Dangers, Strategies},
  publisher    = {Oxford University Press},
  year         = {2014},
}

@book{chang2024pathagi1,
  author       = {Chang, Edward Y.},
  title        = {Multi-{LLM} Agent Collaborative Intelligence; The Path to Artificial General Intelligence, Volume I},
  publisher    = {ACM Books},
  month = {12},
  year         = {2025},
  note = {(First Amazon edition published in March 2024)},
  doi          = {10.1145/3749421},
  isbn         = {979-8-4007-3197-6},
  }

@book{chang2026pathagi2,
  author       = {Chang, Edward Y.},
  title        = {System-2 Reasoning: From Semantic Anchoring to Causal Intelligence: The Path to Artificial General Intelligence, Volume II},
  publisher    = {ACM Books},
  month = {6},
  year         = {2026},
    note = {(First Amazon edition published in February 2026)},
}

@inproceedings{SocraSynthChangCSCI2023,
  author       = {Edward Y Chang},
  title        = {{Examining GPT-4's Capabilities and Enhancement with SocraSynth}},
  booktitle    = {{The $10^{th}$ International Conf. on Comp. Science and Comp. Intelligence}},
  year         = {2023},
  month        = {12},
}

@inproceedings{DikeErisChang2025,
  author       = {Edward Y Chang},
  title        = {{A Checks-and-Balances Framework for Context-Aware Ethical AI Alignment}},
  booktitle    = {{ICML}},
  year         = {2025},
  month        = {7},
}

@inproceedings{chang2026sycophancy,
  ids          = {ChangACL2026},
  author       = {Edward Y. Chang},
  title        = {Diagnosing and Mitigating Sycophancy and Skepticism in {LLM} Causal Judgment},
  booktitle    = {Proceedings of the 64th Annual Meeting of the Association for Computational Linguistics ({ACL})},
  year         = {2026},
}

@misc{chang2026erm,
      title={Epistemic Regret Minimization: Label-Free Causal Critique Beyond Outcome Reward}, 
      author={Edward Y. Chang and Longling Geng},
      year={2026},
      eprint={2602.11675},
      archivePrefix={arXiv},
      primaryClass={cs.AI},
      url={https://arxiv.org/abs/2602.11675}, 
}

@misc{chang2026trivium,
      title={Trivium: Temporal Regret as a First-Class Objective for Causal-Memory Controllers}, 
      author={Edward Y. Chang},
      year={2026},
      eprint={2606.04421},
      archivePrefix={arXiv},
      primaryClass={cs.AI},
      url={https://arxiv.org/abs/2606.04421}, 
}

@book{pearl2009causality,
  author       = {Judea Pearl},
  title        = {Causality: Models, Reasoning, and Inference},
  publisher    = {Cambridge University Press},
  edition      = {2nd},
  year         = {2009},
}

@inproceedings{ouyang2022training,
  ids          = {ouyang2022rlhf, ouyang2022instructgpt},
  author       = {Long Ouyang and Jeffrey Wu and Xu Jiang and Diogo Almeida and Carroll Wainwright and Pamela Mishkin and Chong Zhang and Sandhini Agarwal and Katarina Slama and Alex Ray and John Schulman and Jacob Hilton and Fraser Kelton and Luke Miller and Maddie Simens and Amanda Askell and Peter Welinder and Paul F. Christiano and Jan Leike and Ryan Lowe},
  title        = {Training Language Models to Follow Instructions with Human Feedback},
  booktitle    = {Advances in Neural Information Processing Systems (NeurIPS)},
  volume       = {35},
  pages        = {27730--27744},
  year         = {2022},
  url          = {https://proceedings.neurips.cc/paper_files/paper/2022/hash/b1efde53be364a73914f58805a001731-Abstract.html},
}

@inproceedings{sharma2024sycophancy,
  ids          = {sharma2023sycophancy, sharma2025sycophancy},
  author       = {Mrinank Sharma and Meg Tong and Tomasz Korbak and David Duvenaud and Amanda Askell and Samuel R. Bowman and Newton Cheng and Esin Durmus and Zac Hatfield-Dodds and Scott R. Johnston and Shauna Kravec and Timothy Maxwell and Sam McCandlish and Kamal Ndousse and Oliver Rausch and Nicholas Schiefer and Da Yan and Miranda Zhang and Ethan Perez},
  title        = {Towards Understanding Sycophancy in Language Models},
  booktitle    = {International Conference on Learning Representations},
  year         = {2024},
}

@book{sutton2018rl,
  author       = {Richard S. Sutton and Andrew G. Barto},
  title        = {Reinforcement Learning: An Introduction},
  publisher    = {MIT Press},
  edition      = {2nd},
  year         = {2018},
}

@inproceedings{turpin2023language,
  author       = {Turpin, Miles and Michael, Julian and Perez, Ethan and Bowman, Samuel R.},
  title        = {Language Models Don't Always Say What They Think: Unfaithful Explanations in Chain-of-Thought Prompting},
  booktitle    = {Advances in Neural Information Processing Systems},
  year         = {2023},
  url          = {https://arxiv.org/abs/2305.04388},
  eprint       = {2305.04388},
  archivePrefix= {arXiv},
  primaryClass = {cs.CL},
}

@inproceedings{wei2022chain,
  ids          = {wei2022cot},
  author       = {Wei, Jason and Wang, Xuezhi and Schuurmans, Dale and Bosma, Maarten and Ichter, Brian and Xia, Fei and Chi, Ed and Le, Quoc V. and Zhou, Denny},
  title        = {Chain-of-Thought Prompting Elicits Reasoning in Large Language Models},
  booktitle    = {Advances in Neural Information Processing Systems (NeurIPS)},
  volume       = {35},
  pages        = {24824--24837},
  year         = {2022},
}

@book{suntzu1910art,
  author       = {{Sun Tzu}},
  title        = {The Art of War},
  translator   = {Giles, Lionel},
  publisher    = {Luzac \& Co},
  address      = {London},
  year         = {1910},
  note         = {First annotated English translation; classical treatise on strategic asymmetry and deceptive capability management.},
}

@book{aristotle1999ethics,
  author       = {Aristotle},
  title        = {Nicomachean Ethics},
  translator   = {Irwin, Terence},
  edition      = {2},
  publisher    = {Hackett Publishing},
  address      = {Indianapolis},
  year         = {1999},
  note         = {Original work c. 350 BCE.},
}

@book{russell2019human,
  author       = {Russell, Stuart},
  title        = {Human Compatible: Artificial Intelligence and the Problem of Control},
  publisher    = {Viking},
  address      = {New York},
  year         = {2019},
  isbn         = {978-0525558613},
}

@inproceedings{hadfield2016cirl,
  author       = {Hadfield-Menell, Dylan and Russell, Stuart J. and Abbeel, Pieter and Dragan, Anca},
  title        = {Cooperative Inverse Reinforcement Learning},
  booktitle    = {Advances in Neural Information Processing Systems (NeurIPS)},
  volume       = {29},
  year         = {2016},
}

@book{taleb2012antifragile,
  author       = {Taleb, Nassim Nicholas},
  title        = {Antifragile: Things That Gain from Disorder},
  publisher    = {Random House},
  address      = {New York},
  year         = {2012},
  isbn         = {978-1400067824},
}

@article{urbina2022dualuse,
  author       = {Urbina, Fabio and Lentzos, Filippa and Invernizzi, C{\'e}dric and Ekins, Sean},
  title        = {Dual use of artificial-intelligence-powered drug discovery},
  journal      = {Nature Machine Intelligence},
  volume       = {4},
  number       = {3},
  pages        = {189--191},
  year         = {2022},
  doi          = {10.1038/s42256-022-00465-9},
}

@article{kirilenko2017flashcrash,
  author       = {Kirilenko, Andrei and Kyle, Albert S. and Samadi, Mehrdad and Tuzun, Tugkan},
  title        = {The Flash Crash: High-Frequency Trading in an Electronic Market},
  journal      = {Journal of Finance},
  volume       = {72},
  number       = {3},
  pages        = {967--998},
  year         = {2017},
  doi          = {10.1111/jofi.12498},
}

@inproceedings{ribeiro2020auditing,
  author       = {Ribeiro, Manoel Horta and Ottoni, Raphael and West, Robert and Almeida, Virg{\'i}lio A. F. and Meira Jr., Wagner},
  title        = {Auditing radicalization pathways on {YouTube}},
  booktitle    = {Proceedings of the 2020 Conference on Fairness, Accountability, and Transparency (FAT*)},
  pages        = {131--141},
  year         = {2020},
  doi          = {10.1145/3351095.3372879},
}

@techreport{sec2010flashcrash,
  author       = {{U.S. Securities and Exchange Commission and Commodity Futures Trading Commission}},
  title        = {Findings Regarding the Market Events of {May 6, 2010}},
  institution  = {Joint Advisory Committee on Emerging Regulatory Issues},
  year         = {2010},
  month        = sep,
  note         = {Official regulatory findings on the May~6, 2010 Flash Crash.},
}

@book{rawls1971justice,
  author       = {Rawls, John},
  title        = {A Theory of Justice},
  publisher    = {Harvard University Press},
  address      = {Cambridge, MA},
  year         = {1971},
  isbn         = {978-0674000780},
}

@book{walzer1977just,
  author       = {Walzer, Michael},
  title        = {Just and Unjust Wars: A Moral Argument with Historical Illustrations},
  publisher    = {Basic Books},
  address      = {New York},
  year         = {1977},
  isbn         = {978-0465037070},
}

@article{manheim2019goodhart,
  author       = {Manheim, David and Garrabrant, Scott},
  title        = {Categorizing Variants of {Goodhart}'s Law},
  journal      = {arXiv preprint arXiv:1803.04585},
  year         = {2019},
  url          = {https://arxiv.org/abs/1803.04585},
}

@book{george1999gilgamesh,
  editor    = {George, Andrew},
  title     = {The Epic of Gilgamesh: A New Translation},
  publisher = {Penguin Books},
  address   = {London},
  year      = {1999},
  isbn      = {978-0140447217},
}

@book{major2010huainanzi,
  editor    = {Major, John S. and Queen, Sarah A. and Meyer, Andrew Seth and Roth, Harold D.},
  title     = {The Huainanzi: A Guide to the Theory and Practice of Government in Early Han China},
  publisher = {Columbia University Press},
  address   = {New York},
  year      = {2010},
  isbn      = {978-0231142045},
}

@book{homer1996odyssey,
  author    = {Homer},
  title     = {The Odyssey},
  translator= {Fagles, Robert},
  publisher = {Viking},
  address   = {New York},
  year      = {1996},
  isbn      = {978-0140268867},
}

@book{luo1991threekingdoms,
  author    = {Luo, Guanzhong},
  title     = {Three Kingdoms: A Historical Novel},
  translator= {Roberts, Moss},
  publisher = {University of California Press},
  address   = {Berkeley, CA},
  year      = {1991},
  isbn      = {978-0520224780},
}

@book{beauchamp2019principles,
  author    = {Beauchamp, Tom L. and Childress, James F.},
  title     = {Principles of Biomedical Ethics},
  edition   = {8},
  publisher = {Oxford University Press},
  address   = {New York},
  year      = {2019},
  isbn      = {978-0190640873},
}

@article{near1985whistleblowing,
  author    = {Near, Janet P. and Miceli, Marcia P.},
  title     = {Organizational Dissidence: The Case of Whistle-Blowing},
  journal   = {Journal of Business Ethics},
  volume    = {4},
  number    = {1},
  pages     = {1--16},
  year      = {1985},
  doi       = {10.1007/BF00382668},
}

@article{amodei2016concrete,
  author    = {Amodei, Dario and Olah, Chris and Steinhardt, Jacob and Christiano, Paul and Schulman, John and Man{\'e}, Dan},
  title     = {Concrete Problems in {AI} Safety},
  journal   = {arXiv preprint arXiv:1606.06565},
  year      = {2016},
  url       = {https://arxiv.org/abs/1606.06565},
}

@inproceedings{christiano2017deep,
  author    = {Christiano, Paul F. and Leike, Jan and Brown, Tom B. and Martic, Miljan and Legg, Shane and Amodei, Dario},
  title     = {Deep Reinforcement Learning from Human Preferences},
  booktitle = {Advances in Neural Information Processing Systems (NeurIPS)},
  volume    = {30},
  year      = {2017},
}

@article{bai2022constitutional,
  author    = {Bai, Yuntao and Kadavath, Saurav and Kundu, Sandipan and Askell, Amanda and Kernion, Jackson and Jones, Andy and Chen, Anna and Goldie, Anna and Mirhoseini, Azalia and others},
  title     = {Constitutional {AI}: Harmlessness from {AI} Feedback},
  journal   = {arXiv preprint arXiv:2212.08073},
  year      = {2022},
  url       = {https://arxiv.org/abs/2212.08073},
}

@inproceedings{bender2021stochastic,
  author    = {Bender, Emily M. and Gebru, Timnit and McMillan-Major, Angelina and Shmitchell, Shmargaret},
  title     = {On the Dangers of Stochastic Parrots: Can Language Models Be Too Big?},
  booktitle = {Proceedings of the 2021 ACM Conference on Fairness, Accountability, and Transparency (FAccT)},
  pages     = {610--623},
  publisher = {ACM},
  year      = {2021},
  doi       = {10.1145/3442188.3445922},
}

@misc{haugen2021testimony,
  author       = {Haugen, Frances},
  title        = {Protecting Kids Online: Testimony Before the United States Senate Subcommittee on Consumer Protection, Product Safety, and Data Security},
  howpublished = {United States Senate Committee on Commerce, Science, and Transportation},
  year         = {2021},
  note         = {Testimony on internal Facebook research and platform harms},
}

@article{shih2020global,
  author    = {Shih, Willy C.},
  title     = {Global Supply Chains in a Post-Pandemic World},
  journal   = {Harvard Business Review},
  volume    = {98},
  number    = {5},
  pages     = {82--89},
  year      = {2020},
}

\end{document}